%% file: root.tex
\documentclass[letterpaper, 10 pt, conference]{ieeeconf}  

\IEEEoverridecommandlockouts                              

\input{preamble-commands}

\title{\LARGE \bf
Integration of Fully-Actuated Multirotors into Real-World Applications
}

\author{Azarakhsh Keipour$^{1}$, 
Mohammadreza Mousaei$^{2}$, 
Andrew T Ashley$^{3}$ 
and Sebastian Scherer$^{4}$
\thanks{* This work was supported through NASA Grant Number 80NSSC19C010401.}
\thanks{$^{1,2,4}$ Robotics Institute, Carnegie Mellon University, Pittsburgh, PA
        {\tt\small [keipour, mmousaei, basti]@cmu.edu}}%
\thanks{$^{3}$ University of Pittsburgh, Pittsburgh, PA {\tt\small ata33@pitt.edu}}%
}

\begin{document}

\maketitle
\thispagestyle{empty}
\pagestyle{empty}

\begin{abstract}

\input{abstract}

\end{abstract}


\input{1.Intro}
\input{2.Controller}
\input{3.Attitude-Calculation}
\input{4.Thrust-Calculation}
\input{5.Tests}

\addtolength{\textheight}{-3.3cm}   

\input{6.Conclusion}


\section*{ACKNOWLEDGMENT}

This project became possible due to the support of Near Earth Autonomy (NEA). Also, the authors would like to thank Andrew Saba and Greg Armstrong for their contributions in building the UAVs and performing the experiments.


\bibliographystyle{IEEEtran}
\bibliography{paper-citations.bib}

\end{document}

%% file: preamble-commands.tex
\usepackage{amssymb,amsmath,amsfonts}
\usepackage{graphicx}
\usepackage{subcaption}
\usepackage{tikz}
\usepackage{url}

\usepackage{colortbl}
\usepackage{booktabs}
\usepackage{array}
\usepackage{tabularx}

\newcolumntype{L}[1]{>{\raggedright\let\newline\\\arraybackslash\hspace{0pt}}m{#1}}

\newcommand{\T}{^{\mathstrut\scriptscriptstyle{\top}}} 
\newcommand{\mc}[1]{\ensuremath{\mathcal{#1}}}   

\makeatletter
\DeclareRobustCommand\onedot{\futurelet\@let@token\@onedot}
\def\@onedot{\ifx\@let@token.\else.\null\fi\xspace}

\usepackage[T1]{fontenc}  
\usepackage[b]{esvect}    

\makeatletter
\newlength\xvec@height%
\newlength\xvec@depth%
\newlength\xvec@width%
\newcommand{\xvec}[2][]{%
  \ifmmode%
    \settoheight{\xvec@height}{$#2$}%
    \settodepth{\xvec@depth}{$#2$}%
    \settowidth{\xvec@width}{$#2$}%
  \else%
    \settoheight{\xvec@height}{#2}%
    \settodepth{\xvec@depth}{#2}%
    \settowidth{\xvec@width}{#2}%
  \fi%
  \def\xvec@arg{#1}%
  \def\xvec@dd{:}%
  \def\xvec@d{.}%
  \raisebox{.2ex}{\raisebox{\xvec@height}{\rlap{%
    \kern.05em
    \begin{tikzpicture}[scale=1]
    \pgfsetroundcap
    \draw (.05em,0)--(\xvec@width-.05em,0);
    \draw (\xvec@width-.05em,0)--(\xvec@width-.15em, .075em);
    \draw (\xvec@width-.05em,0)--(\xvec@width-.15em,-.075em);
    \ifx\xvec@arg\xvec@d%
      \fill(\xvec@width*.45,.5ex) circle (.5pt);%
    \else\ifx\xvec@arg\xvec@dd%
      \fill(\xvec@width*.30,.5ex) circle (.5pt);%
      \fill(\xvec@width*.65,.5ex) circle (.5pt);%
    \fi\fi%
    \end{tikzpicture}%
  }}}%
  #2%
}
\makeatother

\newcommand*\matrice[1]{\begin{bmatrix}#1\end{bmatrix}}

\makeatletter
\renewcommand*\env@matrix[1][\arraystretch]{%
  \edef\arraystretch{#1}%
  \hskip -\arraycolsep
  \let\@ifnextchar\new@ifnextchar
  \array{*\c@MaxMatrixCols c}}
\makeatother

\newcommand{\FrameSymbol}{\mc{F}}                       
\renewcommand{\vec}[1]{\xvec[]{#1}}                     
\newcommand{\frm}[1]{\mc{#1}}                           
\newcommand{\Frm}[1]{{{\FrameSymbol}^{\frm{#1}}}}       
\newcommand{\mat}[1]{{\mathbf{#1}}}                     

\newcommand{\axis}[1]{\hat{#1}}

\newcommand{\vecelem}[2]{{_{#2}{#1}}}
\newcommand{\Project}[2]{{_{#2}{#1}}}
\newcommand{\Plane}[2]{{#1 #2}}

\definecolor{commentcolor}{gray}{0.5}
\usepackage{algorithm}
\usepackage{algpseudocode}

\algnewcommand{\LineComment}[1]{\State \textcolor{commentcolor}{\(\triangleright\) #1}}
\algnewcommand{\To}{\textbf{to}}
\algnewcommand{\Break}{\textbf{break}}
\algnewcommand{\Continue}{\textbf{continue}}
\algnewcommand{\IIf}[1]{\State\algorithmicif\ #1\ \algorithmicthen}
\algnewcommand{\EndIIf}{\unskip}
\algnewcommand{\var}[1]{\textit{#1}}
\algnewcommand{\func}[1]{\textsc{#1}}

\newcommand{\FI}{\Frm{I}}
\newcommand{\FB}{\Frm{B}}

\newcommand{\XI}{\axis{X}_{\frm{I}}}
\newcommand{\YI}{\axis{Y}_{\frm{I}}}
\newcommand{\ZI}{\axis{Z}_{\frm{I}}}
\newcommand{\XB}{\axis{X}_{\frm{B}}}
\newcommand{\YB}{\axis{Y}_{\frm{B}}}
\newcommand{\ZB}{\axis{Z}_{\frm{B}}}
\newcommand{\XS}{\axis{X}_{\frm{S}}}
\newcommand{\YS}{\axis{Y}_{\frm{S}}}
\newcommand{\ZS}{\axis{Z}_{\frm{S}}}

\newcommand{\kIv}{\axis{\vec{k}}_{\frm{I}}}

\newcommand{\iSv}{\axis{\vec{i}}_{\frm{S}}}
\newcommand{\jSv}{\axis{\vec{j}}_{\frm{S}}}
\newcommand{\kSv}{\axis{\vec{k}}_{\frm{S}}}

\newcommand{\raxis}{\vec{r}}

\newcommand{\RBI}{\mat{R}_{\frm{BI}}}

\newcommand{\RIS}{\mat{R}_{\frm{IS}}}

\DeclareMathOperator{\cosine}{c}
\DeclareMathOperator{\sine}{s}
\newcommand{\ctheta}{\cosine{\theta}}
\newcommand{\stheta}{\sine{\theta}}
\newcommand{\cphi}{\cosine{\phi}}
\newcommand{\sphi}{\sine{\phi}}
\newcommand{\cpsi}{\cosine{\psi}}
\newcommand{\spsi}{\sine{\psi}}

\newcommand{\spt}{_{sp}}
\newcommand{\des}{_{des}}
\newcommand{\ver}{_{ver}}
\newcommand{\hover}{_{hov}}
\newcommand{\hor}{_{hor}}
\newcommand{\lat}{_{lat}}
\newcommand{\nor}{_{nor}}

\newcommand{\Flmax}{F_{lmax}}
\newcommand{\Flmaxnew}{{F'}_{lmax}}

\newcommand{\Fdes}{\vec{F}\des}
\newcommand{\FdesI}{\vec{F}\des^\frm{I}}

\newcommand{\FdesInew}{\vec{F}\des^{\prime\,\frm{I}}}

\newcommand{\Fspt}{\vec{F}\spt}
\newcommand{\FsptB}{\vec{F}\spt^\frm{B}}

\newcommand{\Fver}{\vec{F}\ver}
\newcommand{\FverB}{\vec{F}\ver^{\frm{B}}}

\newcommand{\Flat}{\vec{F}\lat}

\newcommand{\Fnor}{\vec{F}\nor}

\newcommand{\Fhor}{\vec{F}\hor}
\newcommand{\FhorB}{\vec{F}\hor^{\frm{B}}}

\newcommand{\FhorBnew}{\vec{F}\hor^{\prime\,\frm{B}}}

\newcommand{\FsptBnew}{\vec{F}\spt^{\prime\,\frm{B}}}
\newcommand{\FsptBnewnew}{\vec{F}\spt^{''\,\frm{B}}}

\newcommand{\PlaneB}{{\Plane{\XB}{\YB}}}
\newcommand{\PlaneI}{{\Plane{\XI}{\YI}}}
\newcommand{\PlaneS}{{\Plane{\XS}{\YS}}}

\newcommand{\ProjI}[1]{\Project{#1}{\PlaneI}}

\newcommand{\Fhover}{F\hover}
\newcommand{\FhovB}{\vec{F}\hover^\frm{B}}
\newcommand{\FhovI}{\vec{F}\hover^\frm{I}}

\newcommand{\airlab}{\url{https://theairlab.org/fully-actuated}}

%% file: abstract.tex
The introduction of fully-actuated multirotors has opened the door to new possibilities and more efficient solutions to many real-world applications. However, their integration had been slower than expected, partly due to the need for new tools to take full advantage of these robots. 

As far as we know, all the groups currently working on the fully-actuated multirotors develop new full-pose (6-D) tools and methods to use their robots, which is inefficient, time-consuming, and requires many resources.

We propose a way of bridging the gap between the tools already available for underactuated robots and the new fully-actuated vehicles. The approach can extend the existing underactuated flight controllers to support the fully-actuated robots, or enhance the existing fully-actuated controllers to support existing underactuated flight stacks. We introduce attitude strategies that work with the underactuated controllers, tools, planners and remote control interfaces, all while allowing taking advantage of the full actuation. Moreover, new methods are proposed that can properly handle the limited lateral thrust suffered by many fully-actuated UAV designs. The strategies are lightweight, simple, and allow rapid integration of the available tools with these new vehicles for the fast development of new real-world applications. 

The real experiments on our robots and simulations on several UAV architectures with different underlying controller methods show how these strategies can be utilized to extend existing flight controllers for fully-actuated applications. We have provided the source code for the PX4 firmware enhanced with our proposed methods to showcase an example flight controller for underactuated multirotors that can be modified to seamlessly support fully-actuated vehicles while retaining the rest of the flight stack unchanged.

%% file: 1.Intro.tex
\section{INTRODUCTION} \label{sec:intro}

The past two decades have seen rapid growth in the number of multirotor applications. Traditional multirotors are designed to have co-planar rotors. Although this design choice is simple and maximizes energy efficiency, these UAVs suffer from underactuation (i.e., their rotational motion is unavoidably coupled with their translational motion).

New designs for multirotors have been proposed in recent years that allow fully independent control over both linear and angular motions \cite{Rashad2020,Franchi2019a}. Many other configurations have been proposed to achieve the full actuation while trying to optimize or simplify some aspects of the design or the controller. Some examples of these configurations include modifying the quadrotors for full actuation \cite{Ryll2015}, tetrahedron-shaped hexarotor \cite{Toratani2012}, omni-directional octocopters \cite{Brescianini2016,Brescianini2018a}, semi-coaxial hexarotors \cite{Bershadsky2018}, and hexarotors with additional servos \cite{Ryll2016,Kamel2018a,Kamel2018}. These robots have less energy efficiency and generally have more complex hardware designs than the underactuated UAVs. However, in many cases, the independent control over all six dimensions greatly simplifies the payload and the required approach to a task, ultimately reducing the weight and final costs of the UAV's hardware, software, and development process. Moreover, it makes many new applications possible that were infeasible with underactuated designs. 

    \begin{figure}[t]
        \centering
        \includegraphics[width=\linewidth]{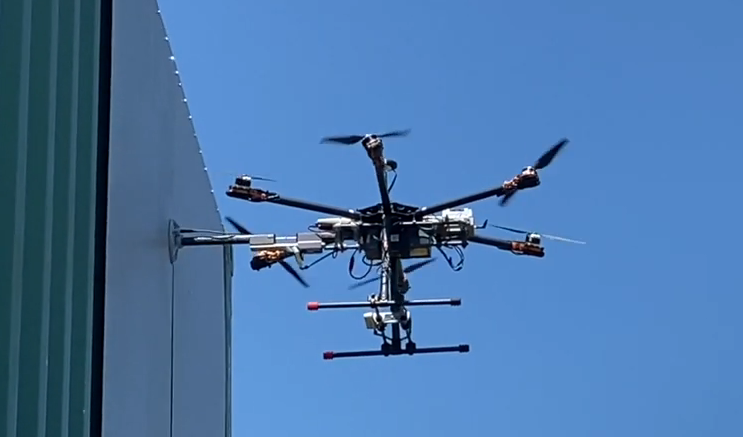}
        \caption{A fully-actuated hexarotor using a controller with the attitude and thrust strategies to make contact with the wall.}
        \label{fig:contact-with-wall}
    \end{figure}

The integration of the new fully-actuated designs into real-life applications had been much slower than the underactuated ones. The added design complexity, the lower energy efficiency, and the additional efforts required to develop new software tools prevent their widespread use in real projects. There have been some efforts to commercialize the fully-actuated multirotor designs (e.g., \cite{Skygauge2020, CyPhy2018, Voliro2020}); however, their success has been minimal so far. 

Over the years, many methods and tools for underactuated multirotors have been developed and are widely available to plan missions and trajectories for different purposes and to control the motion of the robots based on the planned trajectories. The emergence of new architectures has resulted in the introduction of new control methods for the fully-actuated vehicles in different applications, ranging from exact feedback linearization \cite{Rajappa2015, Mehmood2017} to nonlinear model-predictive control \cite{Bicego2019}. However, most of the controllers developed for fully-actuated vehicles require full-pose trajectories and cannot interact with the tools developed for underactuated UAVs. An even more significant challenge is the need for new Remote Control (RC) interfaces that can control the robot in all the six-degrees-of-freedom, making it difficult for the pilots to learn flying or transition to flying these new robots.

On the other hand, many research groups utilize tools such as Simulink to design and generate the code for their new fully-actuated controllers. While these tools can also generate the code required for a full flight control system (such as the state estimation), the generated code is too basic for any tasks performed beyond heavily-controlled lab environments. Moreover, the code cannot be directly integrated with the existing autopilots that already have a more powerful flight stack, slowing down integrating the new robots into real applications even further. 

We have realized that most of the applications do not require a completely free change in orientation and the vehicle's attitude generally needs to follow a set of patterns. We can define a set of \textit{fully-actuated operation modes} for the attitude (we refer to them as \textit{attitude strategies}). The strategies can be switched during the flight to address the different needs of applications. Identifying such a set of strategies allows using the readily-available flight controllers, software tools, and RC interfaces developed for the underactuated UAVs.

Many fully-actuated UAVs, including the most common architectures, such as fixed-pitch hexarotor designs, can generate only limited lateral forces compared to the forces perpendicular to their bodies. Franchi et al. \cite{Franchi2018} have proposed to call these vehicles as \textit{LBF} (vehicles with laterally bounded force), which we will also use in this paper. The operation of the LBF robots requires particular attention to handling their lateral thrust limits. We propose a set of methods (\textit{thrust strategies}) that can handle the lateral thrust limits with minimal changes to the available flight controllers, allowing the use of either 6-D (full-pose) or traditional 4-D planners and motion controllers. 

This paper presents a new way of developing fully-actuated controllers by extending the existing underactuated flight controllers while keeping the entire flight stack without requiring any modifications and allowing to take advantage of the full actuation in the UAV applications. The approach can also be adopted by other fully-actuated controller methods to provide easier integration into existing flight stacks by allowing the controller accept the traditional waypoints and to simplify the development of real-world tasks by allowing easy switching between different strategies. Our proposed modifications to the controller have minimal overhead and can be directly integrated into any autopilot, extending only the Position Controller module.

Our contributions include: 

\begin{itemize}
\renewcommand\labelitemi{--}
    \item Proposing a method to quickly extend an already existing flight controller to work with the fully-actuated robots, enabling the integration of fully-actuated robots with already existing flight stacks and tools available for underactuated multirotors.

    \item Proposing a set of attitude strategies that allow a wide range of applications to use the full actuation for indoor and outdoor environments, and free flight and physical interaction tasks.
    
    \item Proposing a set of thrust strategies to properly handle the limited lateral thrust of the LBF vehicles depending on the application requirements.
    
    \item Providing the source code for the proposed strategies implemented on the PX4 flight controller firmware along with our simulation environment.
\end{itemize}

We show our experiments in Gazebo PX4 SITL and our MATLAB simulator as well as on the real robot to illustrate the proposed strategies and methods.

Section~\ref{sec:controller} discusses the minimum changes required to modify the existing underactuated flight controllers to work with the fully-actuated robots. In Section~\ref{sec:attitude}, we propose the attitude strategies identified to take full advantage of the full actuation using the new controllers. Section~\ref{sec:thrust-setpoint} describes the methods for properly handling the limited lateral thrust in LBF robots. Finally, in Section~\ref{sec:tests}, we present the real and simulated results for the methods proposed in this paper for real-world applications.

%% file: 2.Controller.tex
\section{FLIGHT CONTROLLER ARCHITECTURE} \label{sec:controller}

\begin{figure}[t]
        \centering
        \includegraphics[width=\linewidth]{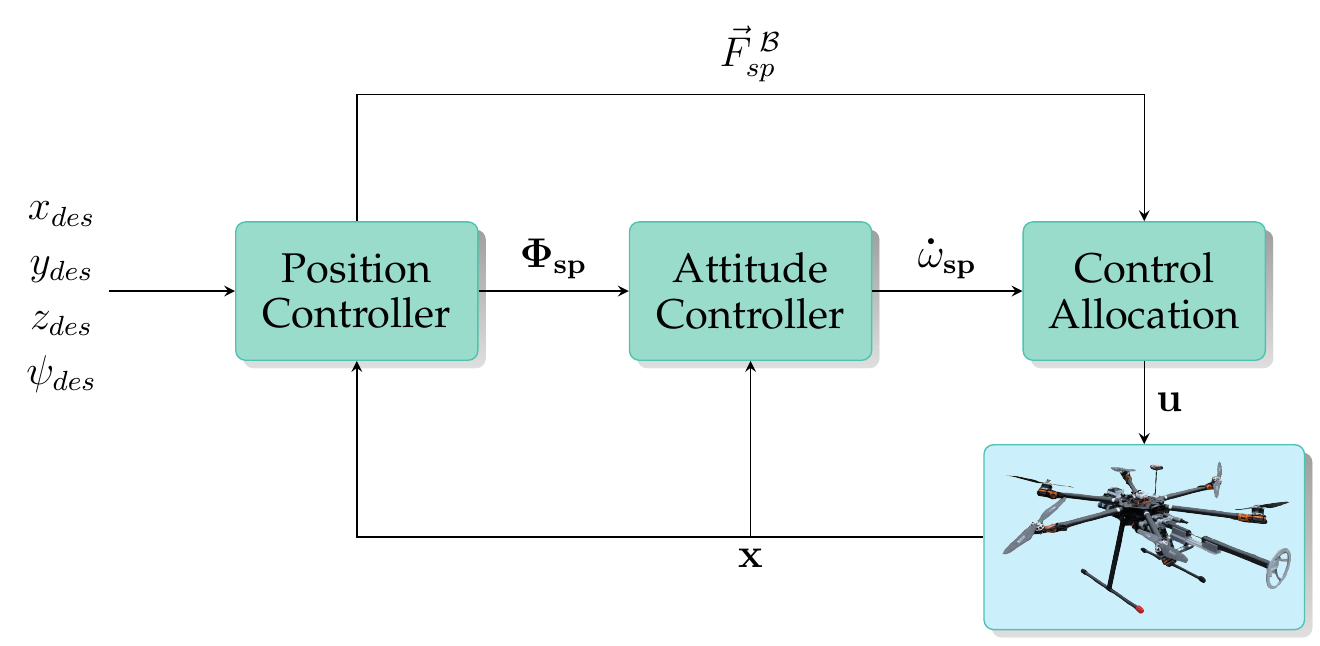}
        
        \caption{A high-level architecture for typical underactuated flight controllers.}
        
        \label{fig:typical-controller-architecture}
    \end{figure}

Figure~\ref{fig:typical-controller-architecture} illustrates the architecture for a typical flight controller for underactuated UAVs. The inputs are normally the desired yaw $\psi\des$, the desired position $p\des = \matrice{x\des & y\des & z\des}\T$, and/or the desired position derivatives. These inputs are internally utilized to generate the full desired attitude $\mat{\Phi\spt}$ (we call it \textit{attitude setpoint}) for its Attitude Controller module and the desired linear acceleration or forces $\Fspt$ (we call it \textit{thrust setpoint}) for its Control Allocation (or Mixer) module. In addition, in Figure~\\ref{fig:typical-controller-architecture} $\mat{x}$ is the state of the UAV, $\mat{u}$ is the input of the robot (e.g., motor PWMs commands), and $\dot{\omega}\spt$ is the output of the Attitude Controller.

\begin{figure}[t]
        \centering
        \includegraphics[width=\linewidth]{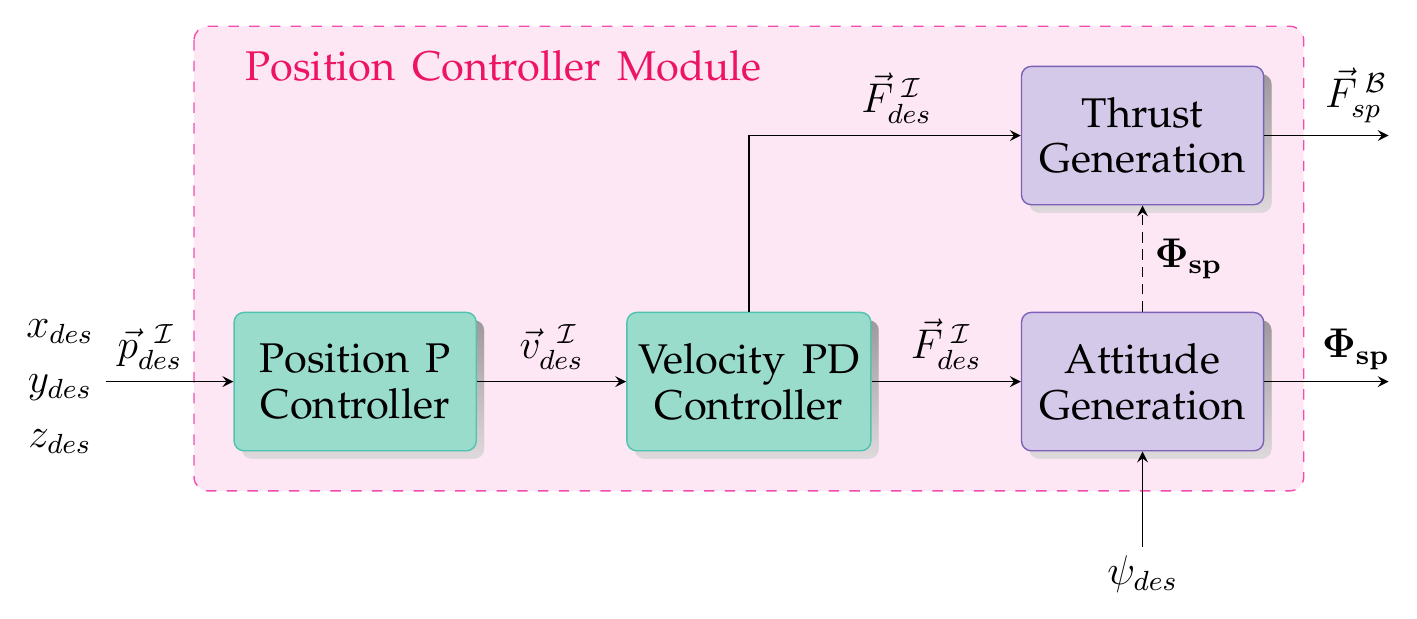}
        
        \caption{The Position Controller module for the PX4 flight controller. The state input $\mat{x}$ is omitted for simplicity.}
        
        \label{fig:position-controller-module}
    \end{figure}

The generation of the attitude and thrust setpoints (i.e., reference attitude and thrust) usually happens within some parent module (e.g., Position Controller). However, here, without the loss of generality, we assume that they are separate submodules. Figure~\ref{fig:position-controller-module} illustrates an example Position Controller module with the thrust and attitude setpoint generation functions depicted as separate submodules.

Some flight controller designs may have slightly different input/output combinations. Many such differences (such as using accelerations instead of forces) do not affect the methods described in this section.

We propose that extending the existing flight controllers is possible by the following minimal set of changes:

\begin{itemize} 
\renewcommand\labelitemi{--}
    \item Modifying or replacing the Control Allocation (or mixer) module to support the new architecture and prioritize angular acceleration over linear acceleration.
    
    \item Extending the Attitude Setpoint Generation function into an Attitude Setpoint Generator module, which allows utilizing the full actuation based on chosen strategies (see Section~\ref{sec:attitude}).
    
    \item Extending (or adding) the Thrust Setpoint Generation function into a Thrust Setpoint Generator module, which manipulates the desired thrust to respect the thrust limits of the fully-actuated LBF vehicles (see Section~\ref{sec:thrust-setpoint}).
\end{itemize}

The described steps can be applied to underactuated controllers, as well as fully-actuated controllers (without the first step), allowing the resulting fully-actuated controller to interact with underactuated flight stacks.

\begin{figure}[t]
        \centering
        \begin{subfigure}[b]{\linewidth}
            \includegraphics[width=\textwidth]{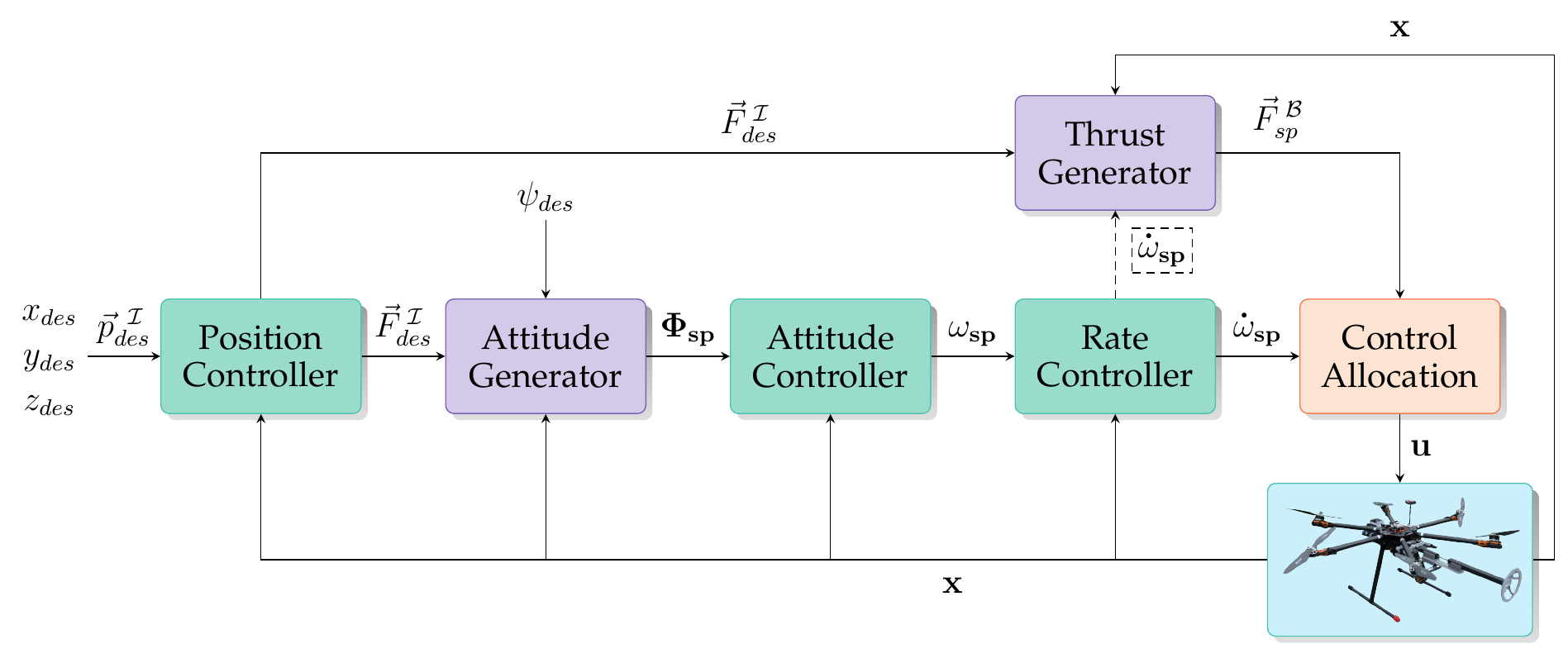}
            \caption{~}
            \label{fig:controller-a}
        \end{subfigure}
        \medskip
        \begin{subfigure}[b]{\linewidth}
            \includegraphics[width=\textwidth]{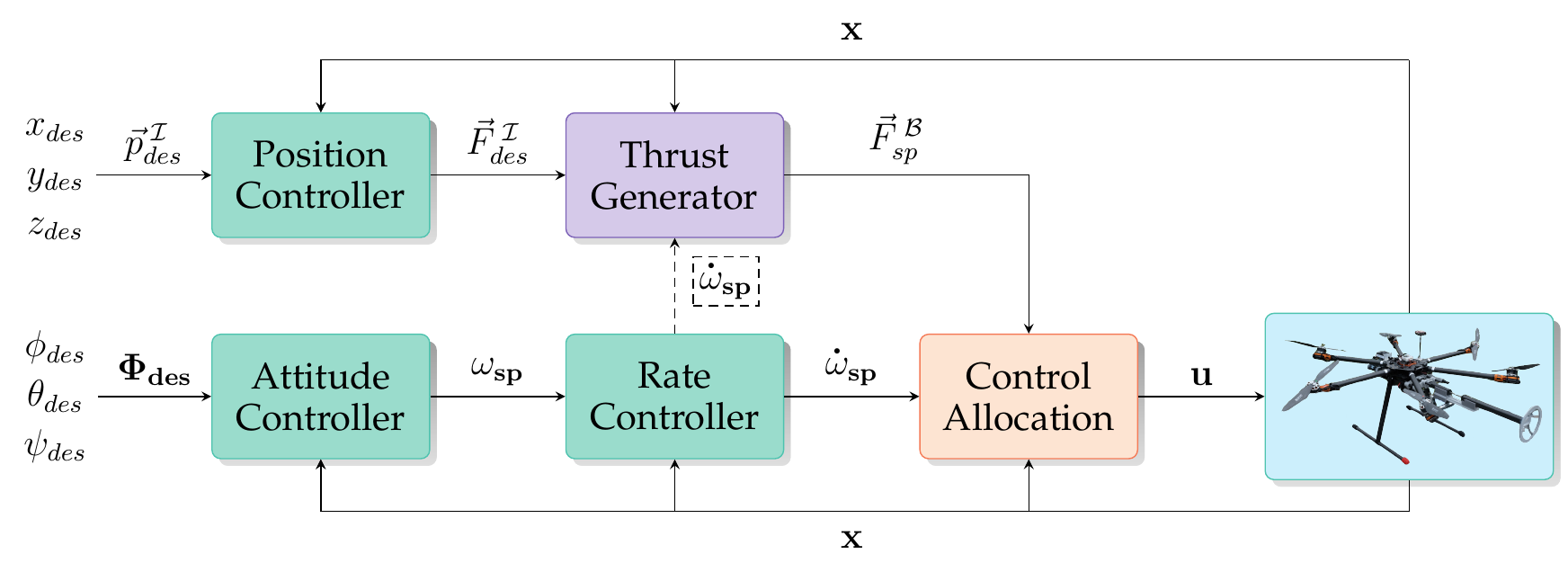}
            \caption{~}
            \label{fig:controller-b}
        \end{subfigure}
        
        \caption{PX4 flight controller extended for fully-actuated multirotors. (a) The input is the desired position (and/or its derivatives) and the desired yaw, requiring an Attitude Generation module to calculate the pitch and roll. (b) The controller input is the desired 6-D pose.}
        
        \label{fig:controller}
    \end{figure}

Figure~\ref{fig:controller} illustrates an example controller extended using the proposed steps. It shows the existing PX4 autopilot architecture \cite{Meier2015} enhanced for fully-actuated multirotors. The controller only modifies the Control Allocation module and extends the Thrust and Attitude Setpoint Generator modules but devises all other parts of the existing controller on PX4 autopilot. Note that the same steps can be applied to any other typical flight controller architecture in a similar manner.

The next two sections explain the Attitude Setpoint Generator and the Thrust Setpoint Generator modules in more detail.

%% file: 3.Attitude-Calculation.tex
\section{REFERENCE ATTITUDE GENERATION} \label{sec:attitude}

\begin{figure}[t]
        \centering
        \includegraphics[width=0.6\linewidth]{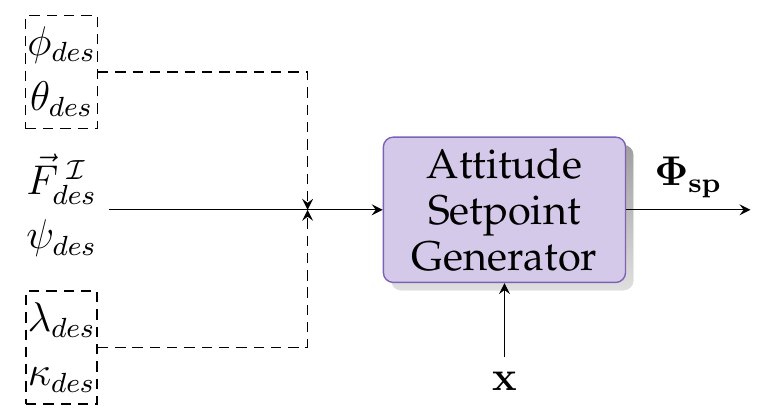}
        \caption{The Attitude Setpoint Generation module.}
        \label{fig:attitude-setpoint-generation-module}
    \end{figure}

The Attitude Setpoint Generator module is illustrated separately in Figure~\ref{fig:attitude-setpoint-generation-module}. Given the desired thrust force in the inertial frame ($\FdesI$), the desired yaw ($\psi\des$), and the UAV's current state $\mat{x}$, the module calculates the full reference attitude ($\mat{\Phi\spt}$) for the Attitude Controller. There are additional strategy-specific inputs (enclosed in dashed boxes), which are not present in the underactuated controllers. These inputs will be explained in their relevant strategies.

We propose a set of five attitude strategies for fully-actuated multirotors that we believe can cover a wide range of applications. These strategies are \textit{zero-tilt}, \textit{full-tilt}, \textit{minimum-tilt}, \textit{fixed-tilt}, and \textit{fixed-attitude}. For each strategy, it is explained how to derive the attitude setpoint in the rotation matrix form. As a reminder, the columns of the rotation matrix are the unit vectors $\iSv$, $\jSv$, and $\kSv$, in the directions of $\XS$, $\YS$ and $\ZS$ axes of the setpoint frame $\Frm{S}$.

We assume that the given desired thrusts have already compensated for the gravity force to achieve the desired acceleration, i.e., the thrust equal to the vehicle's weight is already added to the upward $z$ component of the desired thrust.

Finally, we use the North-East-Down (NED) convention for the inertial frame $\Frm{I}$ and Forward-Right-Down (FRD) convention for the body-fixed $\Frm{B}$ and setpoint $\Frm{S}$ frames.

The rest of this section describes the proposed attitude strategies and the applications where each can be useful.

\subsection{Zero-Tilt Attitude Strategy} \label{sec:attitude:zero-tilt}

Keeping the multirotor's attitude horizontally during the flight can be beneficial for many situations, such as when making precise contact with the vertical surface or capturing a video using an onboard camera, eliminating the need for a gimbal. 

\begin{figure}[t]
        \centering
        \begin{subfigure}[b]{0.49\linewidth}
            \includegraphics[width=\textwidth]{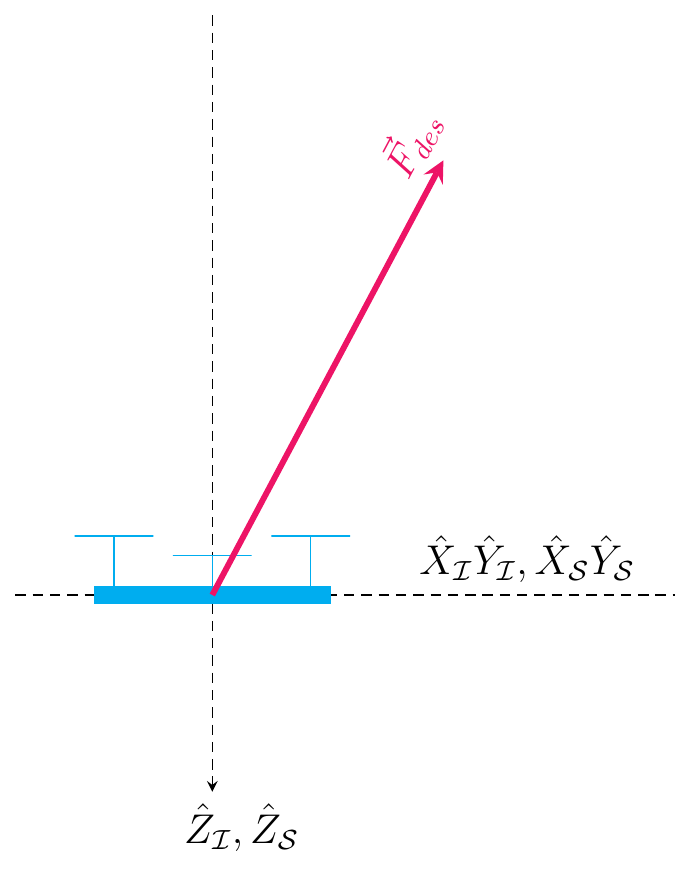}
            \caption{~}
            \label{fig:thrust-strategies-zero-tilt}
        \end{subfigure}
        \hfill
        \begin{subfigure}[b]{0.49\linewidth}
            \includegraphics[width=\textwidth]{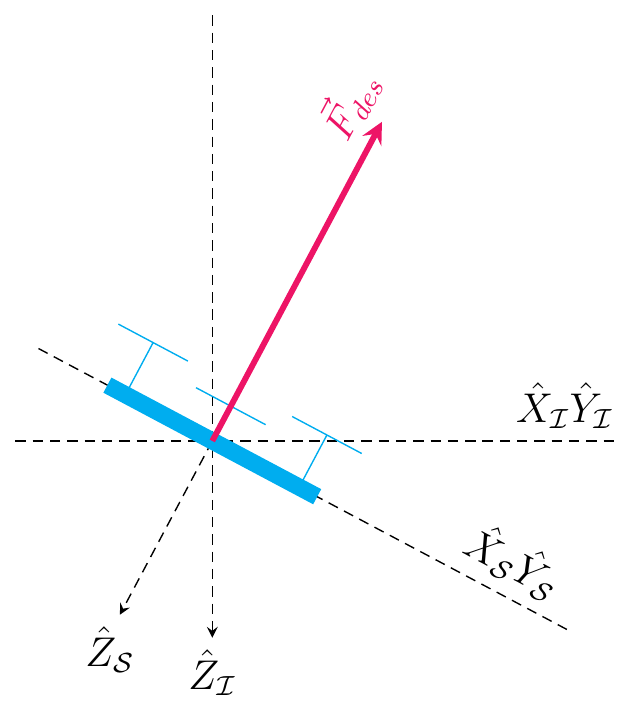}
            \caption{~}
            \label{fig:thrust-strategies-full-tilt}
        \end{subfigure}
        
        \begin{subfigure}[b]{0.49\linewidth}
            \includegraphics[width=\textwidth]{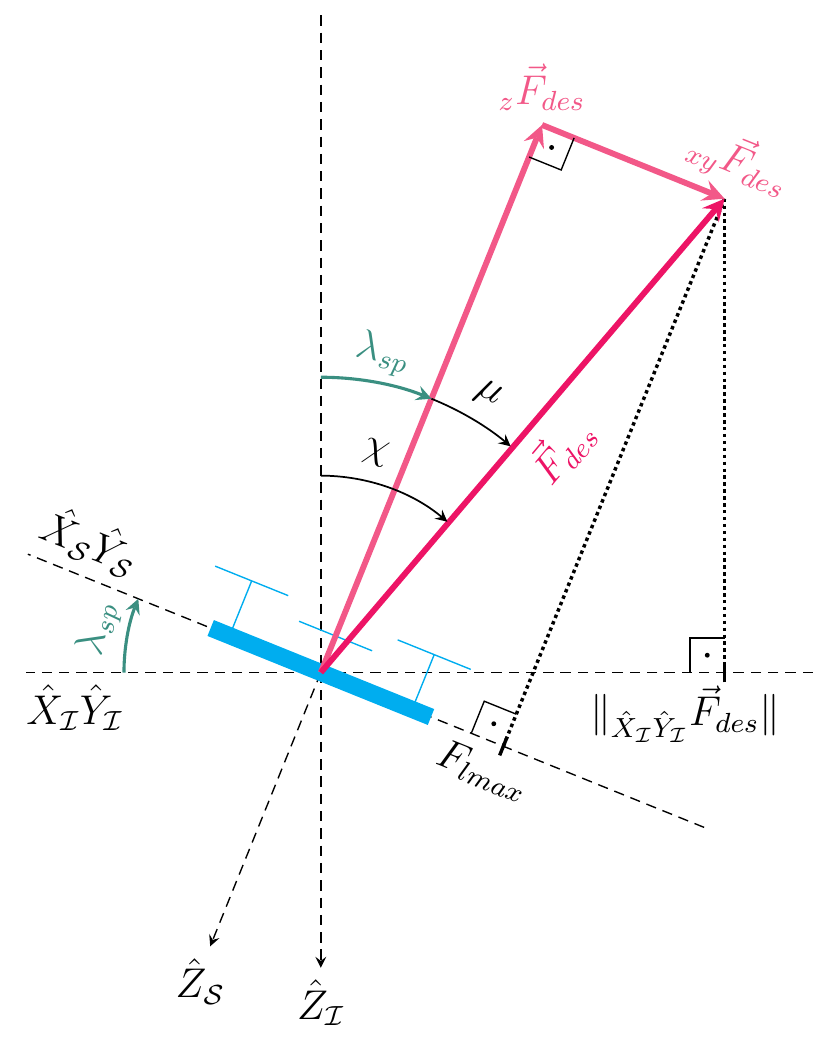}
            \caption{~}
            \label{fig:thrust-strategies-min-tilt}
        \end{subfigure}
        \hfill
        \begin{subfigure}[b]{0.49\linewidth}
            \includegraphics[width=\textwidth]{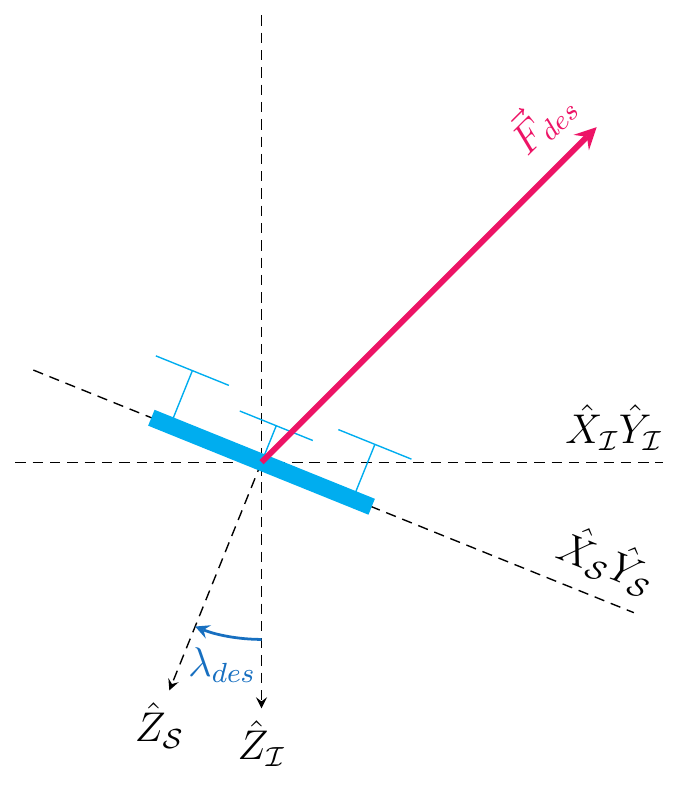}
            \caption{~}
            \label{fig:thrust-strategies-fixed-tilt}
        \end{subfigure}
        
        \caption{Models used for attitude strategies. (a) Zero-tilt. (b) Full-tilt. (c) Minimum-tilt. (d) Fixed-tilt.}
        
        \label{fig:thrust-strategies}
    \end{figure}

Figure~\ref{fig:thrust-strategies-zero-tilt} shows a model used for the zero-tilt attitude generation strategy. 

In this strategy, the direction of $\ZS$ axis for the attitude setpoint is always straight down, the same as $\ZI$ (i.e., $\matrice{0 & 0 & 1}\T$). The $\XS$ axis is in the direction of the input desired yaw $\psi\des$ and $\YS$ axis is perpendicular to it and both lie on the horizontal plane:

    \begin{equation} \label{eq:attitude:zero-tilt-axes}
        \RIS = \begin{bmatrix} 
            \cos \psi\des & -\sin \psi\des & 0\\ 
            \sin \psi\des & \cos \psi\des & 0\\ 
            0 & 0 & 1
            \end{bmatrix}
    \end{equation}

\subsection{Full-Tilt Attitude Strategy} \label{sec:attitude:full-tilt}

With coplanar underactuated multirotor designs, the only way for the robot to achieve the input desired thrust is to tilt so that the direction of the desired thrust is normal to the plane of the rotors. This strategy is the most useful to oppose the external forces and disturbances and reduce the total energy consumption when control of the tilt is not crucial in a given task.

Figure~\ref{fig:thrust-strategies-full-tilt} shows a model that demonstrates the full-tilt attitude strategy. $\FdesI$ should be normal to the $\PlaneS$ plane and in the opposite direction of $\ZS$ axis. Therefore, we have $\kSv = -{\FdesI}/{\left\| \FdesI \right\|}$.

If an imaginary unit vector in the desired yaw direction on the $\PlaneI$ plane is rotated for $+90^\circ$ around $\ZI$ axis, it will fall on the plane made from axes $\YS$ and $\ZS$. Thus, $\iSv = \matrice{-\sin\psi\des & \cos\psi\des & 0}\T \times \kSv$. Finally, we have $\jSv = \kSv \times \iSv$.

\subsection{Minimum-Tilt Attitude Strategy} \label{sec:attitude:min-tilt}

The zero-tilt strategy described in Section~\ref{sec:attitude:zero-tilt} requires a lateral thrust input that is less than the maximum lateral thrust limit $\Flmax$ in LBF robots. In some applications, keeping the tilt as close to zero as possible is desirable, but achieving the desired thrusts and accelerations is more important than maintaining the zero-tilt attitude. An example scenario is filming a highly dynamic target or in high-gust winds. This strategy, first described in \cite{Mehmood2017}, minimize the robot's tilt by using up the lateral thrust. Figure~\ref{fig:thrust-strategies-min-tilt} helps demonstrating how the minimum-tilt attitude is calculated. 

If the horizontal component of $\FdesI$ is less than the maximum lateral thrust $\Flmax$, then the zero-tilt attitude strategy can be used to calculate the attitude setpoint. Otherwise, all the available thrust on the lateral plane is used first, then the remaining required thrust determines the required tilt $\lambda\spt$, roll $\phi\spt$ and pitch $\theta\spt$. To calculate the tilt $\lambda\spt$ we have:

    \begin{equation} \label{eq:attitude:min-tilt-calculation}
        \lambda\spt = \underbrace{ \arcsin \left(\frac{\left\| \ProjI{\Fdes} \right\|}{\left\| \Fdes \right \|}\right)}_\chi
        - \underbrace{ \arcsin\left( \frac{\Flmax}{\left\| \Fdes \right \|} \right) }_\mu
    \end{equation}

\noindent where $\chi$ and $\mu$ angles are as illustrated in Figure~\ref{fig:thrust-strategies-min-tilt}, and $\ProjI{\Fdes}$ is the projection of the desired thrust on the horizontal plane.

The axis of rotation $\raxis$ is perpendicular to the plane consisting of the desired thrust $\Fdes$ and the inertial $\ZI$ axis. Hence, it can be calculated as $\raxis = (\Fdes \times \kIv)/{\left\| \Fdes \times \kIv \right \|}$.
    
Finally, by rotating the $\kIv$ unit vector around $\raxis$ using the Rodrigues' rotation formula we have:

    \begin{equation} \label{eq:attitude:min-tilt-zdes}
    \begin{split}
        \kSv = &(1 - \cos\lambda\spt) \ (\raxis \cdot \kIv) \ \raxis + \kIv \ \cos\lambda\spt\\
        &+ (\raxis \times \kIv) \ \sin\lambda\spt
    \end{split}
    \end{equation}

The unit vectors for $\XS$ and $\YS$ axes can be calculated similarly to the full-tilt strategy.

\subsection{Fixed-Tilt Attitude Strategy} \label{sec:attitude:fixed-tilt}

Some applications require keeping a specific tilt angle for the multirotor. An example scenario is flying in constant wind when keeping a fixed tilt against the wind independent of the yaw and the movement direction is desirable to increase the thrust that remains after opposing the wind. 

Figure~\ref{fig:thrust-strategies-fixed-tilt} shows a model demonstrating the axes and angles used in the calculation of the fixed-tilt attitude strategy.

The strategy requires two additional inputs $\lambda\des$ and $\kappa\des$ to the system, representing the angle of the desired tilt and the direction of the tilt, respectively (see Figure~\ref{fig:attitude-setpoint-generation-module}). 
We assume here that the desired direction $\kappa\des$ is given in the inertial frame (i.e., with respect to the north direction). 

The axis of rotation $\raxis$ to tilt the robot is perpendicular to $\ZI$ axis and the projection of the vector pointing in the direction of tilt on the $\PlaneI$ plane. Hence, it can be computed as $\raxis = \matrice{\cos\kappa\des & \sin\kappa\des & 0}\T \times \kIv$.
    
Finally, the unit vector for the $\ZS$ axis is calculated using Equation~\ref{eq:attitude:min-tilt-zdes}, and the $\XS$ and $\YS$ axes are calculated similarly to the full-tilt strategy.

\subsection{Fixed-Attitude Strategy} \label{sec:attitude:fixed-attitude}

In practice, some applications require keeping specific attitude angles without explicitly including the roll and pitch angles in the trajectory. An example scenario is during the robot's contact with the wall, when keeping a constant orientation may help control the end-effector's pose and wrench. 

The strategy requires two new inputs: the desired roll $\phi\des$ and the desired pitch $\theta\des$ (see Figure~\ref{fig:attitude-setpoint-generation-module}). The rotation matrix can be directly calculated from the given Euler angles:

    \begin{equation} \label{eq:attitude:rotation-matrix-from-euler-angles}
    \begin{split}
        &\RIS = \\
        &\begin{bmatrix} 
            \ctheta\cpsi & \sphi\stheta\cpsi - \cphi\spsi & \cphi\stheta\cpsi + \sphi\spsi\\
            \ctheta\spsi & \sphi\stheta\spsi + \cphi\cpsi & \cphi\stheta\spsi - \sphi\cpsi\\
            -\stheta & \sphi\ctheta & \cphi\ctheta
        \end{bmatrix}
    \end{split}
    \end{equation}

\noindent where $c$ and $s$ are shorthand for $\cos$ and $\sin$ functions, respectively, and $\phi$, $\theta$, $\psi$ are used for $\phi\des$, $\theta\des$ and $\psi\des$.

Note that in the fixed-attitude strategy, when the yaw changes, the robot changes its tilt direction to keep the given roll and pitch, which is different from the fixed-tilt strategy, where the tilt direction is fixed independent of the given yaw. 

\subsection{Discussion}

Each of the five attitude strategies can be used at different parts of an application depending on the given task.

When keeping a specific attitude is not critical, the full-tilt strategy is the preferred method, as it consumes the least energy and has maximum robustness to external disturbances such as winds and gusts. 

The zero-tilt strategy is a special case of both fixed-tilt and fixed-attitude strategies. However, in practice, it is used often enough that having an easy way to switch to this strategy without any required input parameters can simplify the workflow. 

Full-tilt and minimum-tilt strategies are the only methods to achieve the desired input acceleration (if at all feasible) when required (e.g., following an agile subject). However, it comes at the cost of not having any guarantees on the attitude of the robot. Note that both methods achieve the same tilt direction, with a smaller angle for minimum-tilt strategy.

The proposed set of strategies can be extended with new methods when there is a need for a specific application. This can be achieved simply by adding the additional method to the Attitude Setpoint Generator module. 

%% file: 4.Thrust-Calculation.tex
\section{THRUST SETPOINT GENERATION} \label{sec:thrust-setpoint}

The output of the Thrust Generator module in Figure~\ref{fig:controller} is expressed in $\FB$ frame, while its input is in $\FI$ frame. With an unlimited available thrust, the input thrust can be simply rotated from $\FI$ to $\FB$ (i.e., $\FdesI$ can be projected on the current body-fixed axes) to compute the thrust setpoint: $\FsptB = \RBI \cdot \FdesI$.

In practice, the available thrust is not unlimited. In particular, fully-actuated LBF multirotors have even more limited lateral thrust than the thrust normal to their body. If the element of $\Fdes$ on the body $\PlaneB$ plane is larger than the maximum lateral thrust ($\Flmax$), some motors will saturate, and the system may lose its stability.

Franchi et al. \cite{Franchi2018} have proposed a method to handle the limited thrust of LBF robots. Their work targets a specific flight control architecture and ultimately requires 6-D planners and motion controllers to work. However, we propose thrust strategies that can handle the lateral thrust limits with minimal changes to the existing flight controllers, allowing both full-pose and traditional underactuated planners and motion controllers. 

First, we should define the terminology:

\textit{Lateral thrust} ($\Flat$): The component of the thrust on the body-fixed $\PlaneB$ plane.

\textit{Normal thrust} ($\Fnor$): The component of the thrust on the body-fixed $\ZB$ axis.

\textit{Horizontal thrust} ($\Fhor$): The component of the thrust on the inertial $\PlaneI$ plane.

\textit{Vertical thrust} ($\Fver$): The component of the thrust on the inertial $\ZI$ axis. 

\textit{Hover thrust} ($\Fhover$): The vertical thrust required to keep the UAV hovering when no force other than gravity is acting on it. 

\textit{Maximum lateral thrust} ($\Flmax$): The maximum achievable force on the lateral plane ($\PlaneB$) of the UAV in the direction of the desired force. Here we assume that $\Flmax$ is a constant scalar.

Given $\FdesI$, two cases can happen:

\noindent \textit{Case 1.} $\| \Flat \| \leq \Flmax$: Generating $\FdesI$ is feasible.

\noindent \textit{Case 2.} $\| \Flat \| > \Flmax$: Generating $\FdesI$ is not feasible. 

We describe two different sets of strategies with different objectives for handling the latter case. Each method has its strengths and can be used depending on the situation.

\subsection*{Method 1: Keeping the Desired Vertical Thrust}

Merely cutting the lateral input to $\Flmax$ bounds can result in losing a portion of the vertical thrust, leading to altitude tracking errors and, in extreme cases, catastrophic results. Therefore, an important objective when handling the lateral thrust can be keeping the vertical desired thrust. 

First, both the vertical and horizontal components of $\FdesI$ (i.e., $\Fver$ and $\Fhor$) are projected on $\FB$ separately to compute the partial thrust setpoint vectors $\FverB$ and $\FhorB$.

We assume that the lateral component of $\FverB$ is not greater than $\Flmax$, otherwise the vertical desired thrust cannot be achieved. The assumption is reasonable if the robot's tilt is limited (the limits can be set based on the dynamics of the multirotor). After consuming the lateral thrust required for the vertical component, the remaining available lateral thrust $\Flmaxnew$ can be used to bound the horizontal thrust $\FhorB$: 

    \begin{equation} \label{eq:thrust:lateral-thrust-output-bounding-horizontal}
        \vec{F}\hor^{'\frm{B}} =
        \frac{\Flmax}{\sqrt{\left(\vecelem{F\hor^\frm{B}}{x}\right)^2 + \left(\vecelem{F\hor^\frm{B}}{y}\right)^2}} 
        \ \FhorB
    \end{equation}

Finally, the thrust setpoint is $\FsptBnew = \FverB + \FhorBnew$.

A different method is explained in \cite{Mehmood2017}, which directly constructs the thrust setpoint by utilizing the knowledge that the maximum lateral thrust is already being fully consumed.

\subsection*{Method 2: Keeping the Acceleration Directions}

Another method for handling the lateral thrust limits reduces both the vertical and horizontal thrusts simultaneously but ensures that the accelerations' direction is maintained. 

Knowing the hover thrust $\Fhover$, if the $z$ component of the input desired thrust is larger than $\Fhover$, then the hover thrust vector $\FhovI = \matrice{0 & 0 & \Fhover}\T$ is rotated to the current body-fixed frame to obtain the baseline for zero acceleration ($\FhovB$).

The lateral thrust consumed by the hover thrust is used to obtain the new lateral thrust limit $\Flmaxnew$. The rest of the desired thrust ($\FdesInew = \FdesI - \FhovI$) is rotated to $\FB$ to get $\FsptBnew$. Then the lateral thrust of $\FsptBnew$ is bounded to $\Flmaxnew$ to get $\FsptBnewnew$. Finally, $\FsptBnewnew$ is combined with $\FhovB$ to calculate the feasible thrust setpoint $\FsptB$.

\

Method~1 guarantees that the output thrust setpoint has the same vertical component as the input desired thrust if it is feasible. However, it may cause a severe reduction in the remaining lateral thrust for horizontal motion when the vertical thrust command is large, or the tilt angle is high. Method~2 fixes the issue by reducing the vertical component to ensure that some thrust is left for the horizontal motion.

%% file: 5.Tests.tex
\section{EXPERIMENTS AND RESULTS} \label{sec:tests}

We have tested the proposed methods in simulation and on a real robot. This section explains our hardware and software setups and the experiments.

\subsection{Robot's Hardware and Software} \label{sec:tests:robot}

We built fixed-tilt hexarotors with all rotors rotated sideways alternatively for 30 and -30 degrees, similar to the design described in \cite{Mehmood2017}. 

The robots' main frame is Tarot T960 with KDE-3510XF-475 motors, KDE-UAS35HVC ESCs, and 14-inch propellers. They are equipped with a mRo Pixracer autopilot hardware, an Nvidia Jetson TX2 onboard computer, a u-Blox Neo-M8N GPS module, Futaba T10J transmitter/receiver, and a 900~MHz radio for communication with the Ground Control Station computer. 

The onboard computer runs Linux Ubuntu 18.04 (Bionic Beaver) with Robot Operating System (ROS) Melodic Morenia. Depending on the task at hand, different software packages run to plan and control the missions and trajectories.

\begin{figure}[t]
        \centering
        \includegraphics[width=0.6\linewidth]{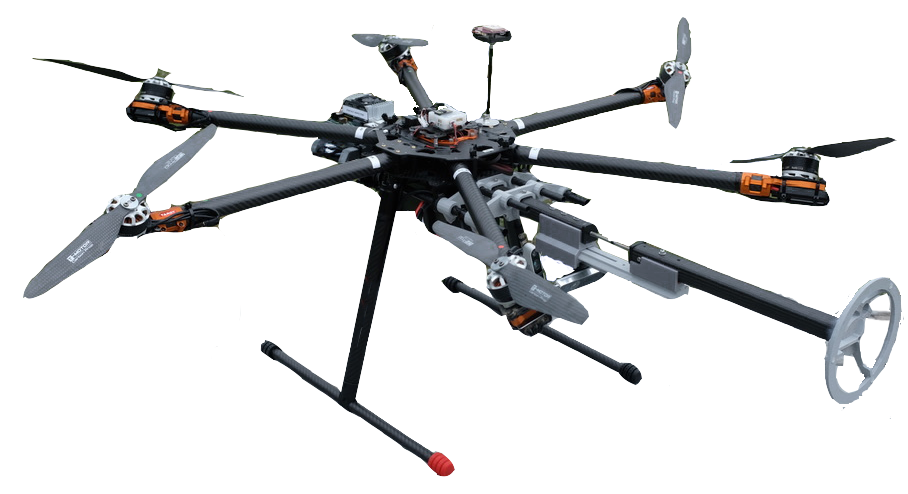}
        \caption{A fixed-tilt UAV used in our experiments.}
        \label{fig:hardware}
    \end{figure}

Figure~\ref{fig:hardware} shows one of the UAVs we have used for our experiments.

We enhanced several recent versions of PX4 firmware (including v1.10 and v1.11) to support fully-actuated vehicles and implemented the methods presented in this paper (see Figure~\ref{fig:controller}). The source code for the firmware and the documentation can be accessed from \airlab.

\subsection{Simulation} \label{sec:tests:simulation}

Two simulation environments were devised for testing the new fully-actuated UAV development and the methods described in this paper: the Gazebo simulator with PX4 SITL and a MATLAB simulator.

\begin{figure}[t]
        \centering
        \begin{subfigure}[b]{0.49\linewidth}
            \includegraphics[width=\textwidth, height=25mm]{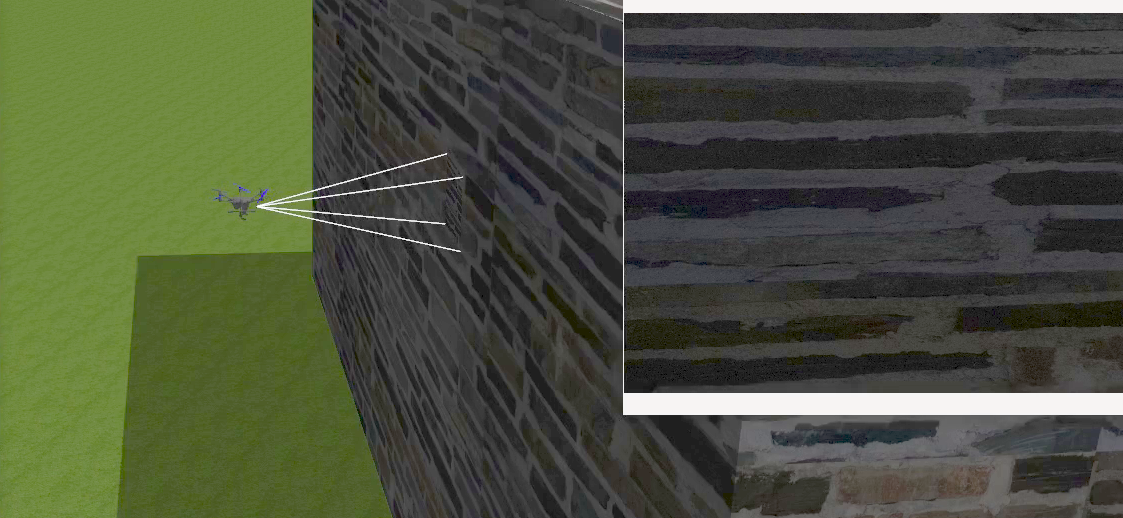}
            \caption{~}
            \label{fig:sim-gazebo-sample}
        \end{subfigure}
        \hfill
        \begin{subfigure}[b]{0.49\linewidth}
            \includegraphics[width=\textwidth, height=25mm]{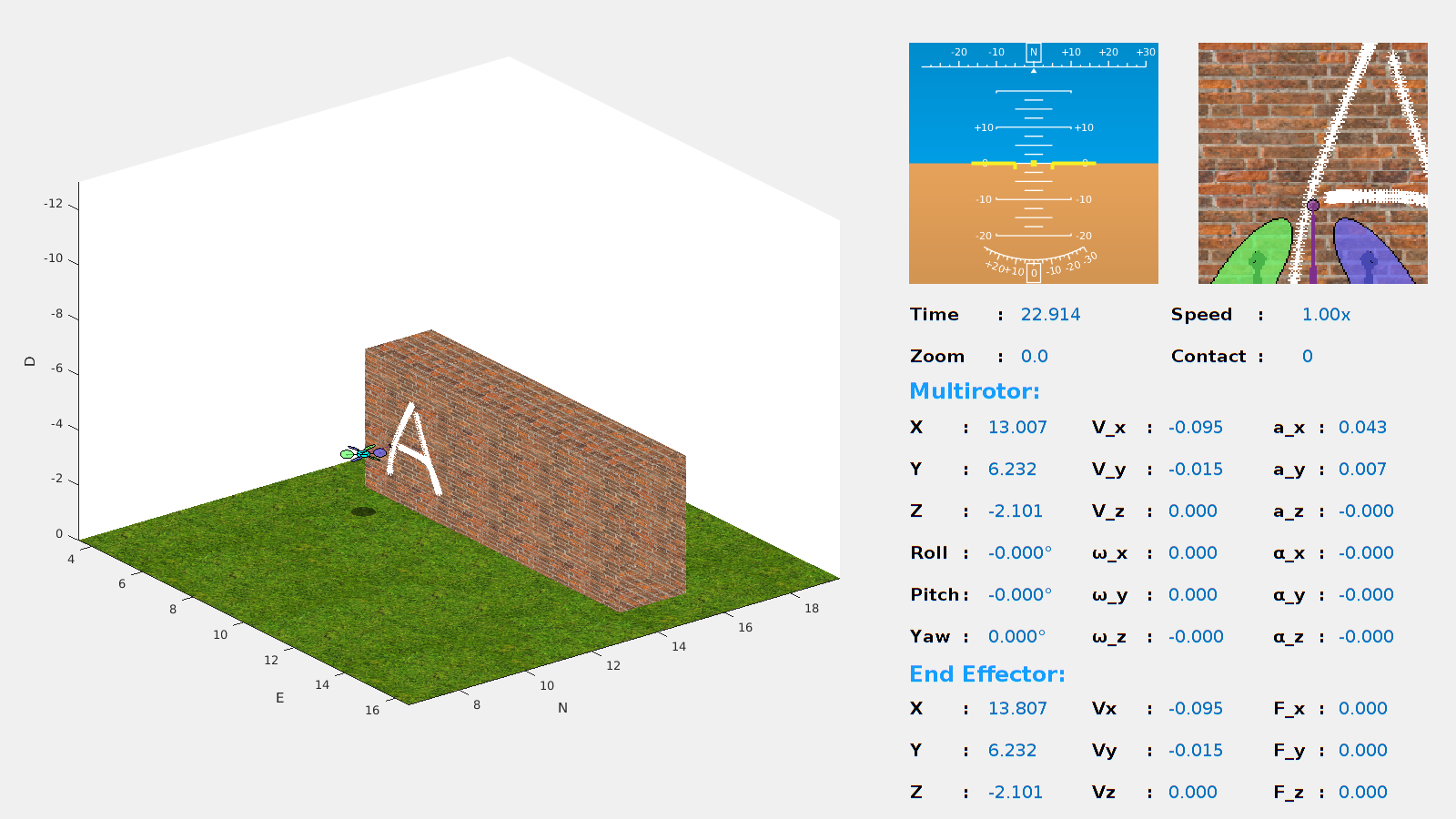}
            \caption{~}
            \label{fig:sim-matlab-sample}
        \end{subfigure}
        
        \caption{Screenshots from the simulation environments showing our fully-actuated hexarotor while running the controller and strategies proposed in this paper. (a) Gazebo with PX4 SITL making unmodeled contact with the wall. (b) MATLAB simulator writing on the wall.}
        \label{fig:sim-sample}
    \end{figure}

The developed MATLAB simulator can be used to develop, analyze, visualize, and test different architectures and methods quickly. The screenshot of the simulator in Figure~\ref{fig:sim-matlab-sample} shows the model of our UAV using the proposed methods combined with force control to write a text on the wall.

The Gazebo simulator uses the SITL simulation provided by the PX4 firmware, which we have enhanced to support fully-actuated vehicles and the proposed methods and include our fully-actuated hexarotor model. It is mostly used for testing the code developed for the real robots in simulation before performing real experiments. Our PX4 firmware code's documentation explains how to run this simulation with the new fully-actuated UAVs and switch between the strategies described in this paper.

To compare the performance of the attitude strategies, we used a trajectory with two waypoints to simulate all five attitude strategies in our MATLAB simulator. The starting point is inertial zero with zero attitude. The waypoints are $\matrice{2 & 2 & -4 & 0}\T$ and $\matrice{2 & 6 & -3 & 30}\T$, respectively. The waypoints' elements are $x$, $y$, $z$, and $yaw$ in degrees. The hexarotor with fixed-pitch arms of our project is used at all the trials.

\begin{figure}[t]
        \centering
        \begin{subfigure}[b]{0.49\linewidth}
            \includegraphics[width=\textwidth]{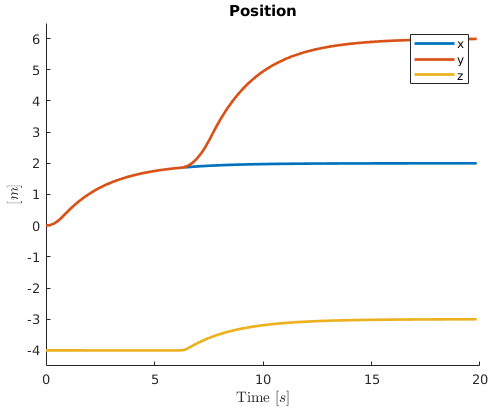}
            \caption{~}
            \label{fig:thrust-strategies-sim-pos}
        \end{subfigure}
        \hfill
        \begin{subfigure}[b]{0.49\linewidth}
            \includegraphics[width=\textwidth]{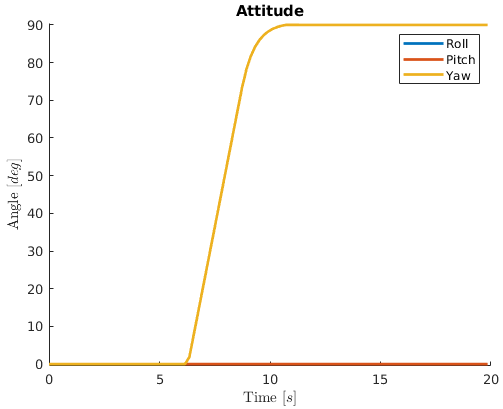}
            \caption{~}
            \label{fig:thrust-strategies-sim-zero-tilt}
        \end{subfigure}
        
        \begin{subfigure}[b]{0.49\linewidth}
            \includegraphics[width=\textwidth]{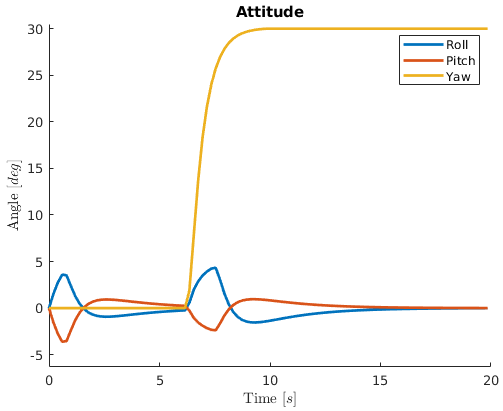}
            \caption{~}
            \label{fig:thrust-strategies-sim-full-tilt}
        \end{subfigure}
        \hfill
        \begin{subfigure}[b]{0.49\linewidth}
            \includegraphics[width=\textwidth]{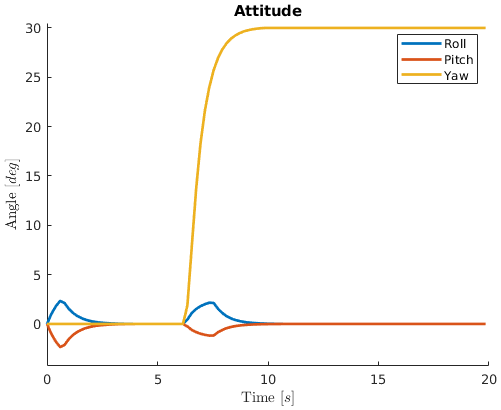}
            \caption{~}
            \label{fig:thrust-strategies-sim-min-tilt}
        \end{subfigure}
        
        \begin{subfigure}[b]{0.49\linewidth}
            \includegraphics[width=\textwidth]{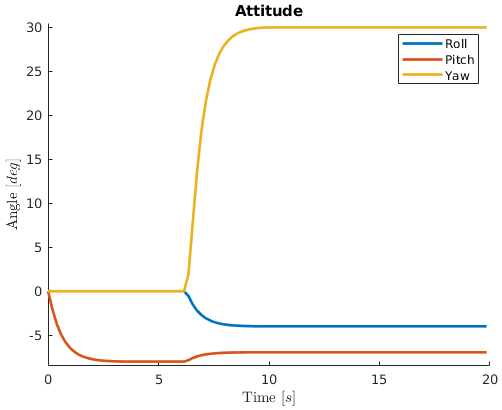}
            \caption{~}
            \label{fig:thrust-strategies-sim-fixed-tilt}
        \end{subfigure}
        \hfill
        \begin{subfigure}[b]{0.49\linewidth}
            \includegraphics[width=\textwidth]{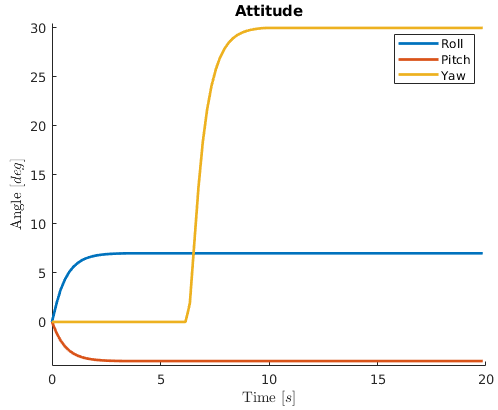}
            \caption{~}
            \label{fig:thrust-strategies-sim-fixed-att}
        \end{subfigure}
        
        \caption{Simulation results of the robot flying a two-waypoint trajectory. (a) Position of the robot. (b) Attitude using the zero-tilt strategy. (c) Attitude using the full-tilt strategy. (d) Attitude using the minimum-tilt strategy. (e) Attitude using the fixed-tilt strategy. (f) Attitude using the fixed-attitude strategy.}
        
        \label{fig:thrust-strategies-sim-results}
    \end{figure}

Figure~\ref{fig:thrust-strategies-sim-pos} shows the path followed by the UAV and Figures~\ref{fig:thrust-strategies-sim-results}(b-f) illustrate the attitudes of the robot when flying the given path. 

It can be observed that the tilt of the robot with the minimum-tilt strategy is less than when using the full-tilt strategy, but both are in the same direction. On the other hand, in the fixed-tilt strategy, the robot has to change its roll and pitch to keep its tilt direction when the yaw is changing. Finally, as expected, the robot's roll and pitch with zero-tilt and fixed-attitude strategies are independent of the given position and yaw setpoint.

\subsection{Real-World Experiments} \label{sec:tests:tests}

\begin{figure}[t]
        \centering
        \begin{subfigure}[b]{0.49\linewidth}
            \includegraphics[width=\textwidth]{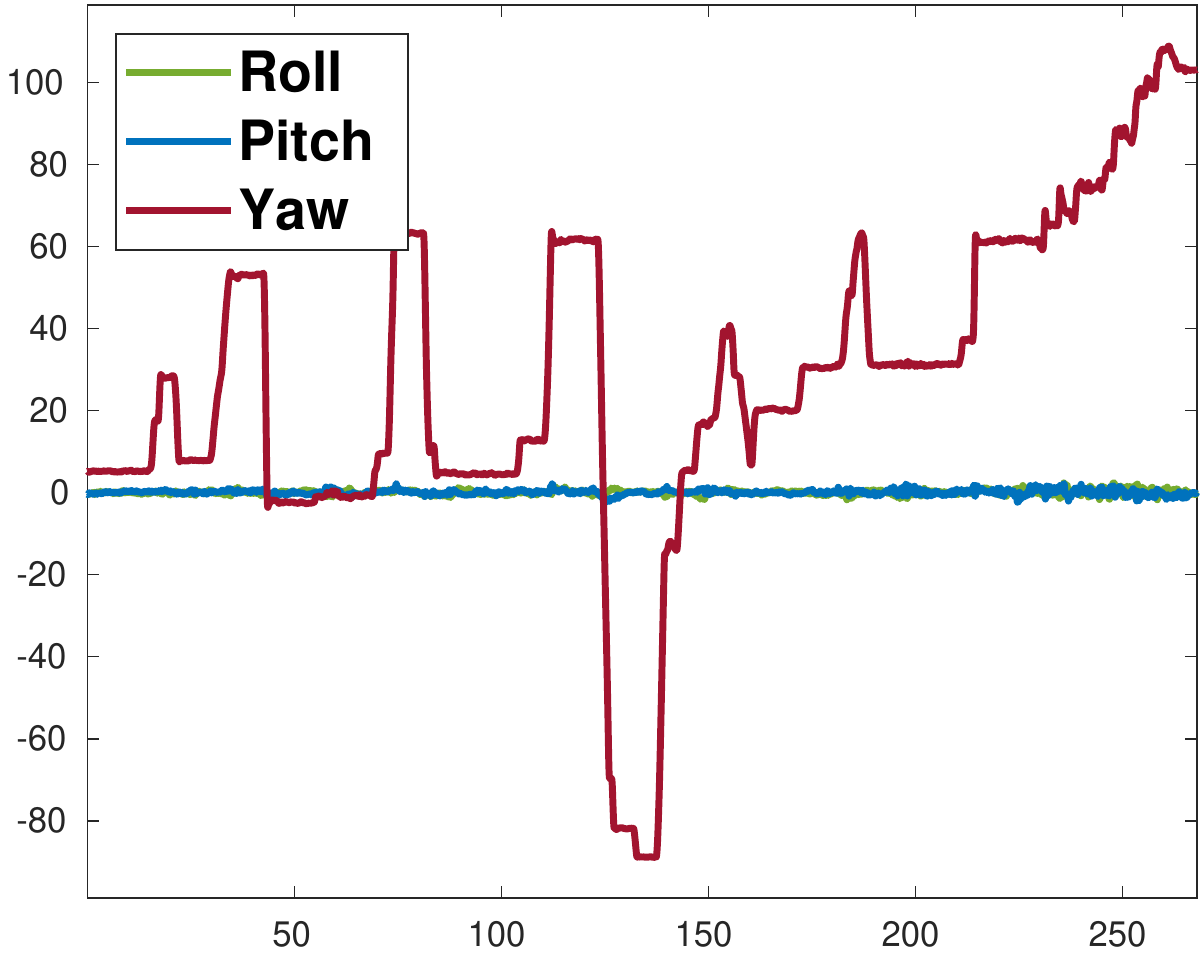}
            \caption{~}
            \label{fig:zero-tilt-results-rpy}
        \end{subfigure}
        \hfill
        \begin{subfigure}[b]{0.49\linewidth}
            \includegraphics[width=\textwidth]{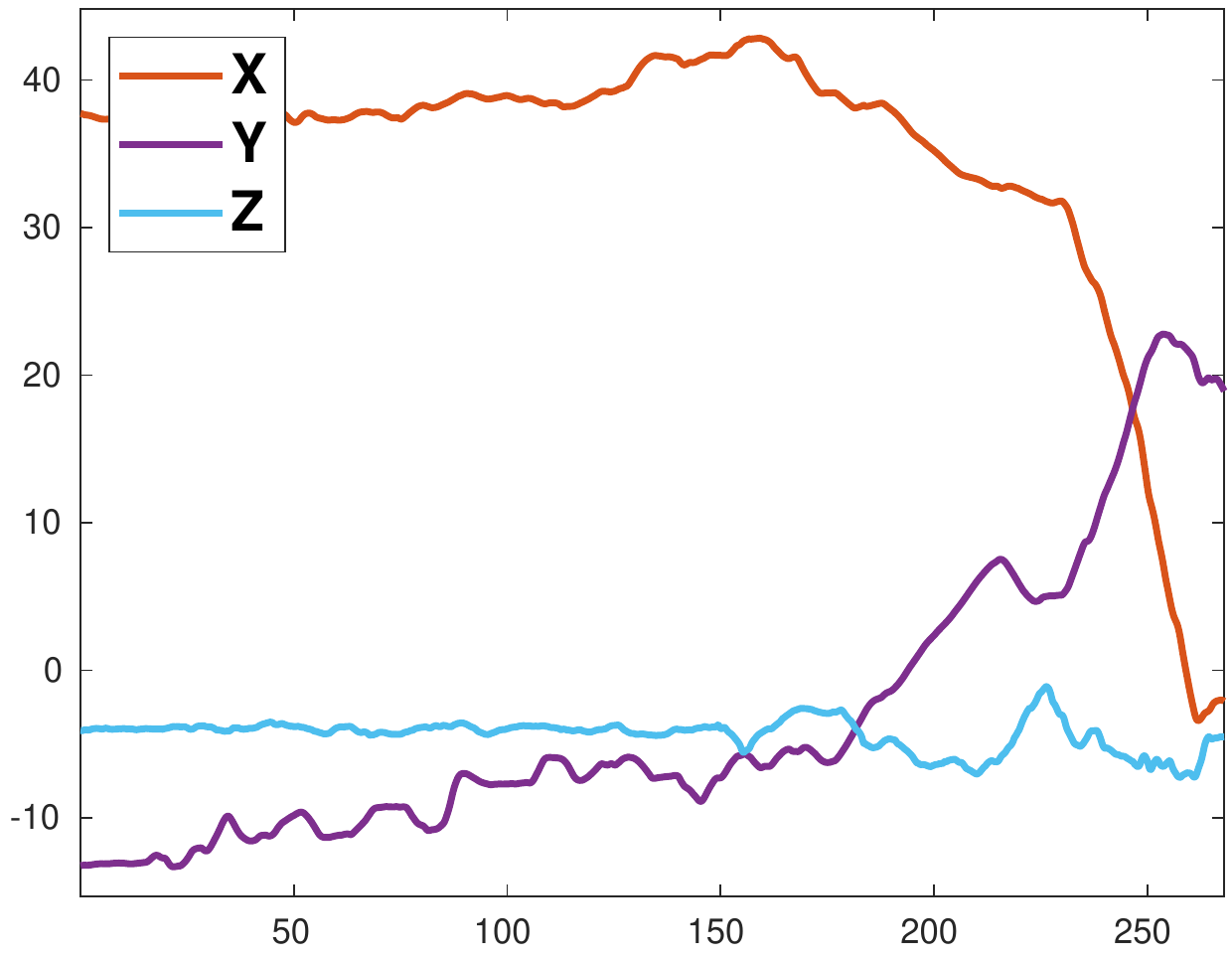}
            \caption{~}
            \label{fig:zero-tilt-results-pos}
        \end{subfigure}
        
        \caption{An outdoor flight segment with the zero-tilt strategy in the presence of winds and gusts. The yaw changes aggressively, and the multirotor is flying around while the roll and pitch stay close to zero. }
        
        \label{fig:zero-tilt-results}
    \end{figure}

We have performed tens of flights with our fully-actuated hexarotor platform running the proposed controller and strategies. The experiments include real flights in the wind and physical contact with the environment. Figure~\ref{fig:contact-with-wall} shows our UAV running the zero-tilt strategy during contact with a wall to measure the properties of the contact point using an ultrasonic sensor. The contact is unmodeled, but the UAV can keep its zero-tilt attitude and reject the disturbance. Figure~\ref{fig:zero-tilt-results} shows the robot keeping its zero-tilt attitude while aggressively flying and turning. 

\begin{figure}[t]
        \centering
        \begin{subfigure}[b]{0.49\linewidth}
            \includegraphics[width=\textwidth]{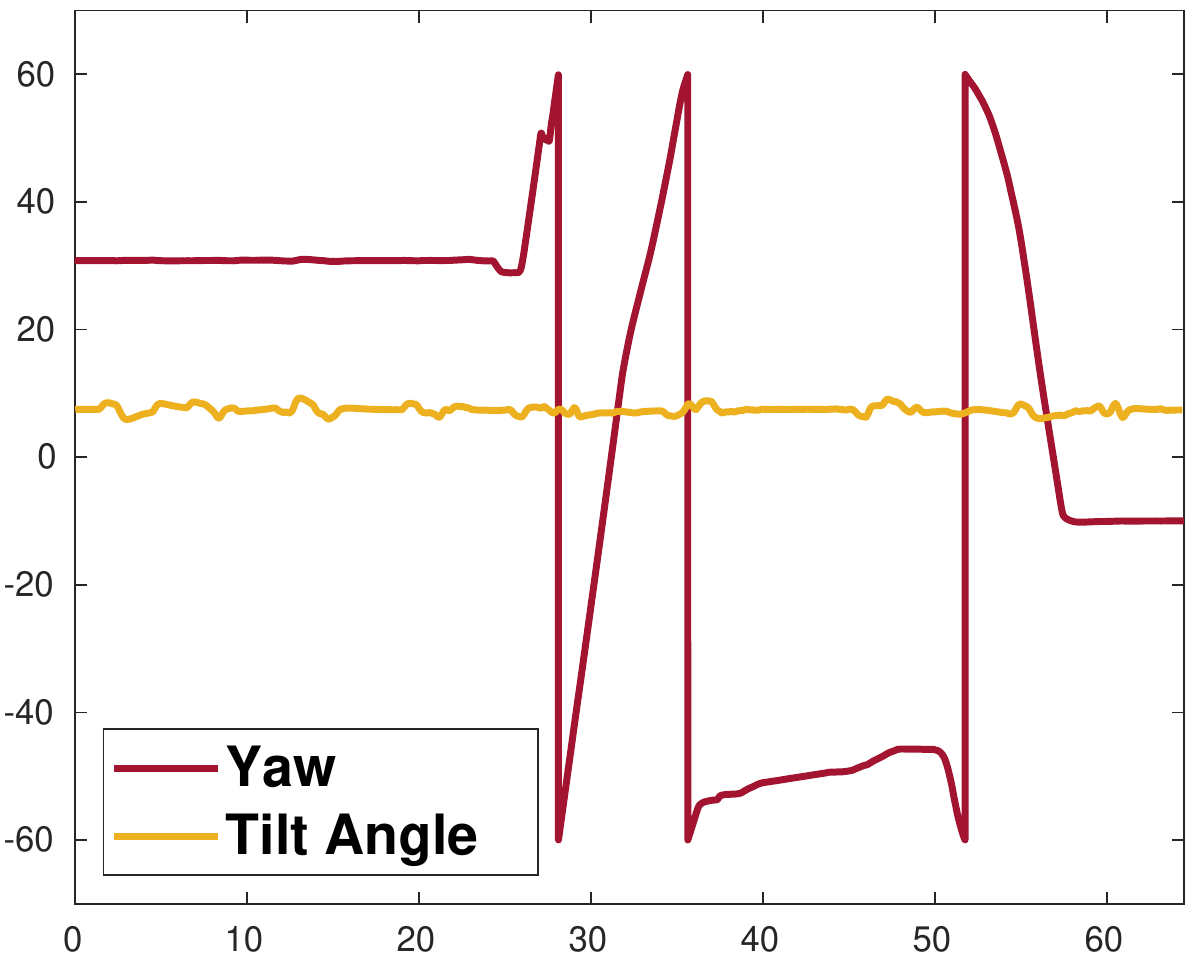}
            \caption{~}
            \label{fig:fixed-tilt-results-y}
        \end{subfigure}
        \hfill
        \begin{subfigure}[b]{0.49\linewidth}
            \includegraphics[width=\textwidth]{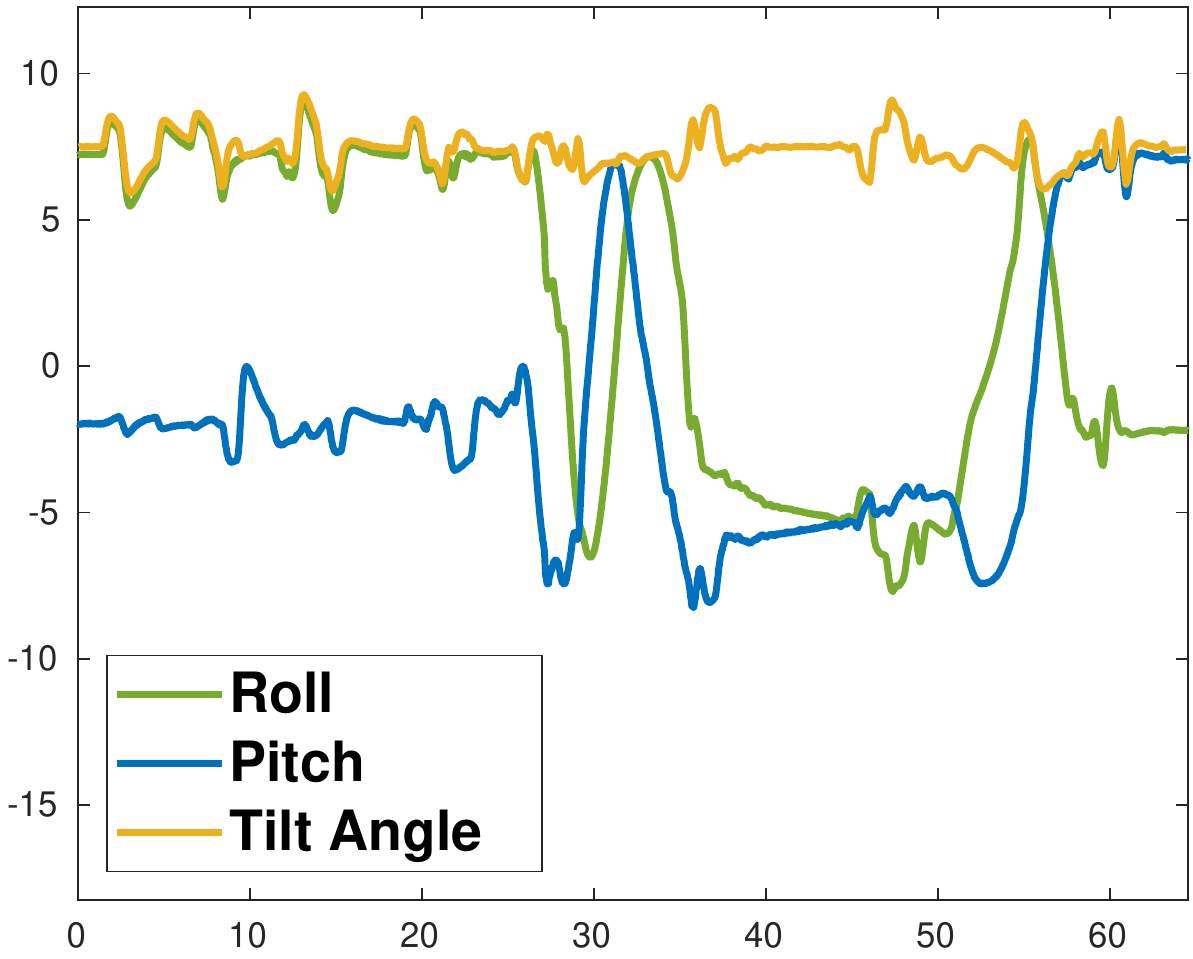}
            \caption{~}
            \label{fig:fixed-tilt-results-rp}
        \end{subfigure}
        
        \caption{An outdoor flight segment with the fixed-tilt strategy in the presence of winds and gusts. The tilt is locked around 7.5 degrees, while the yaw changes aggressively. The right plot shows that while the tilt angle is constant, the roll and pitch change when the yaw changes. The yaw is scaled by $1/3$ in the plot.}
        
        \label{fig:fixed-tilt-results}
    \end{figure}

Figure~\ref{fig:fixed-tilt-results} plots another flight segment flying with the fixed-tilt attitude strategy in a strong and gusty wind. The figure illustrates the tilt angle staying almost constant while the multirotor performs aggressive motions.
    
The video provided with this paper shows the free flights and physical interaction of our robot, as well as the simulation for the attitude and thrust strategies working with several fully-actuated hardware architectures.

\ 

The experiments illustrate that integrating the proposed strategies can successfully extend an underactuated flight controller to allow a fully-actuated robot to be used with available underactuated flight stack and tools. Note that the proposed strategies only generate setpoints for the underlying controller to track, and the tracking precision of the controller depends on the underlying method used and not as much on the added strategies.

%% file: 6.Conclusion.tex
\section{CONCLUSIONS} \label{sec:conclusion}

In this paper, we described a quick way to enhance the available underactuated multirotor flight controllers to support the fully-actuated multirotors while retaining the ability to interact with all the available tools and interfaces for both the traditional underactuated and the new fully-actuated robots. The approach can also be utilized to enhance an existing fully-actuated controller allowing it to be used with underactuated flight stacks and tools. 

We proposed a set of attitude strategies that can accept the same trajectories as the underactuated robots but fully take advantage of the 6-D actuation. We also proposed a set of thrust strategies with different objectives that can enhance the robots' performance with limited lateral thrust force (LBF multirotors). 

Finally, we illustrated our methods working in a simulator with several different architectures and on real robots, keeping the intended full pose both in free flight and during physical interaction with the environment. The real-world experiments were performed with a robot equipped with a popular open-source autopilot firmware minimally modified to work with fully-actuated UAVs and enhanced with our proposed attitude and thrust strategies. 

We hope that the described methods help the researchers and the industry accelerate the integration of fully-actuated robots into real-world applications.